\documentclass[runningheads]{llncs}
\usepackage{graphicx}

\usepackage{tikz}
\usepackage{comment}
\usepackage{amsmath,amssymb} 
\usepackage{color}

\usepackage[accsupp]{axessibility}  

\usepackage{booktabs}
\usepackage{pifont}

\usepackage[width=122mm,left=12mm,paperwidth=146mm,height=193mm,top=12mm,paperheight=217mm]{geometry}
\usepackage[pagebackref,breaklinks,colorlinks]{hyperref}
\usepackage[capitalize]{cleveref}

\newcommand{\cmark}{\ding{51}}

\begin{document}
\pagestyle{headings}
\mainmatter
\def\ECCVSubNumber{100}  

\title{Point-Voxel Adaptive Feature Abstraction for Robust Point Cloud Classification} 

\titlerunning{PV-Ada for Robust Point Cloud Classification}
%
\author{Lifa Zhu\inst{1} \and
Changwei Lin\inst{1} \and
Chen Zheng\inst{2} \and
Ninghua Yang\inst{1}}
\authorrunning{L. Zhu et al.}
%
\institute{DeepGlint, Beijing, China \and
China University of Mining and Technology, Beijing, China \\
\email{zhulf0804@gmail.com}\\
\email{\{changweilin,ninghuayang\}@deepglint.com}\\
\email{zc824435664@163.com}}
\maketitle

\begin{abstract}
Great progress has been made in point cloud classification with learning-based methods. However, complex scene and sensor inaccuracy in real-world application make point cloud data suffer from corruptions, such as occlusion, noise and outliers. In this work, we propose Point-Voxel based Adaptive (PV-Ada) feature abstraction for robust point cloud classification under various corruptions. Specifically, the proposed framework iteratively voxelize the point cloud and extract point-voxel feature with shared local encoding and Transformer. Then, adaptive max-pooling is proposed to robustly aggregate the point cloud feature for classification. Experiments on ModelNet-C dataset demonstrate that PV-Ada outperforms the state-of-the-art methods. In particular, we rank the $2^{nd}$ place in ModelNet-C classification track of PointCloud-C Challenge 2022, with Overall Accuracy (OA) being 0.865. Code will be available at \href{https://github.com/zhulf0804/PV-Ada}{https://github.com/zhulf0804/PV-Ada}.
\end{abstract}

\section{Introduction}
\label{sec:intro}

Point cloud classification, as one of the important 3D tasks, has achieved remarkable progress\cite{qi2017pointnet,qi2017pointnet++,thomas2019kpconv,li2018pointcnn,wang2019dynamic,guo2021pct,zhao2021point,ma2022rethinking,pang2022masked}. However, the state-of-the-art methods are less robust in 3D point cloud recognition under corruptions or attack\cite{ren2022modelnet-c,sun2022benchmarking,sun2022pointdp,perez20223deformrs}. In this report, we mainly describe our method and experiments on ModelNet-C point cloud classification hosted in PointCloud-C Challenge 2022\footnote{https://pointcloud-c.github.io/competition.html}, which benchmarks point cloud robustness analysis under corruptions\cite{ren2022modelnet-c}.

ModelNet-C is the first systematically-designed test-suite for point cloud classification under corruptions\cite{ren2022modelnet-c}. It extends the validation set of ModelNet40 with the proposed atomic corruptions, including “Add Global”, “Add Local”, “Drop Global”, “Drop Local”, “Rotate”, “Scale” and “Jitter”. Moreover, each atomic corruption is performed with 5 severity levels. It's noted that each real-world corruption can be broken down into a combination of the atomic corruptions.

In terms of the corruptions in ModelNet-C data, we propose Point-Voxel Adaptive (PV-Ada) feature abstraction for robust point cloud classification. PV-Ada is proposed based on the following two observations. First, the rough contour may benefit point cloud classification under corruptions, instead of paying attention to detailed structures. It inspires us to involve both voxel-level and point-level feature for point cloud classification. Second, not all all points in point cloud are equal for individual feature representation, such as plane points and outlier points. So we propose to learn point weight in an end-to-end manner, thus conducting adaptive pooling for robust feature abstraction. 

Experiments on ModelNet-C and ModelNet demonstrate the effectiveness of PV-Ada. Our method outperforms 
the state-of-the-art published models and achieves 0.884 and 0.865 OA on the public and private ModelNet-C test set, respectively.

\section{Method}
\label{sec:method}

A point cloud is represented as a set $\mathcal{X}=\{\mathbf x_i \in \mathbb R^3\}_{i = 1, 2, \cdots, N}$,  where $N$ is the point number in $\mathcal{X}$. For the convenience of following discussion, point cloud is denoted by matrix $\mathbf X \in \mathbb R^{N \times 3}$. Inputted point cloud $\mathbf X$, PV-Ada outputs $C$ scores for all the $C$ candidate categories. The overall architecture of PV-Ada is depicted in \cref{fig:pv-ada}.

\subsection{Point-voxel encoder}
\label{subsec:encoder}

Point-voxel as input and voxelization operation have been widely used in point cloud detection, segmentation and classification\cite{shi2020pv,shi2021pv,liu2022pvnas,wu2022pv,li2021voxel,liu2019point,thomas2019kpconv,choy2019fully,choy20194d}, which shows the effectiveness and efficiency of voxelization. Our presented point-voxel encoder consists of three modules: local encoding, Transformer and pyramid feature interaction.

\paragraph{Local encoding} Local encoding groups $k$ nearest neighbors for each point (voxel), then learns $D$ dimensional point (voxel) feature through convolution and non-linear operations. Take $\mathbf{X} \in \mathbb R^{N \times 3}$ as an example, local encoding outputs its local desciptor $\mathbf{F}_{\mathbf X}^e \in \mathbb R^{N \times D}$. 

\paragraph{Transformer} We adopt PCT\cite{guo2021pct} to enhance the point feature for point cloud. Take $\mathbf{X}$ as an example, it first further encode point feature $\mathbf{F}_{\mathbf X}^e$ with MLP and non-linear operations, generating $D$ dimensional features $\mathbf{F}_{\mathbf X}^m \in \mathbb R^{N \times D}$. Then four offset-attention are introduced to obtain features $(\mathbf{F}_{\mathbf X}^{oa_{1}}, \mathbf{F}_{\mathbf X}^{oa_{2}}, \mathbf{F}_{\mathbf X}^{oa_{3}}, \mathbf{F}_{\mathbf X}^{oa_{4}}) \in \mathbb R^{N \times D}$. Each is generated through a vanilla attention and a offset-based residual structure:
\begin{equation}
\begin{aligned}
  \mathbf{F}_{\mathbf X}^{oa_{0}} &= \mathbf{F}_{\mathbf X}^m, \\
  \mathbf{F}_{\mathbf X}^{va_{i}} &= \text{VanillaAttention}(\mathbf{F}_{\mathbf X}^{oa_{i-1}}), \quad i = 1, 2, 3, 4, \\
  \mathbf{F}_{\mathbf X}^{oa_{i}} &= \mathbf{F}_{\mathbf X}^{oa_{i-1}} + f_i(\mathbf{F}_{\mathbf X}^{oa_{i-1}} - \mathbf{F}_{\mathbf X}^{va_{i}}),
  \label{eq:oa_attention}
\end{aligned}
\end{equation}
where $f_i(\cdot)$ is implemented using Conv1d, normalization and ReLU. 

Finally, the feature $\mathbf{F}_{\mathbf X}^t \in \mathbb R^{N \times {4D}}$ outputted from Transformer is calculated through:
\begin{equation}
\begin{aligned}
  \mathbf{F}_{\mathbf X}^t &= f_5(\text{cat}([\mathbf{F}_{\mathbf X}^e, \mathbf{F}_{\mathbf X}^{oa_{1}}, \mathbf{F}_{\mathbf X}^{oa_{2}}, \mathbf{F}_{\mathbf X}^{oa_{3}}, \mathbf{F}_{\mathbf X}^{oa_{4}}])),
\end{aligned}
\end{equation}
where $\text{cat}(\cdot)$ is concatenation, $f_5(\cdot)$ is implemented using Conv1d, normalization and LeakyReLU.

\begin{figure*}[!t]
\centering
\includegraphics[width=.98\textwidth]{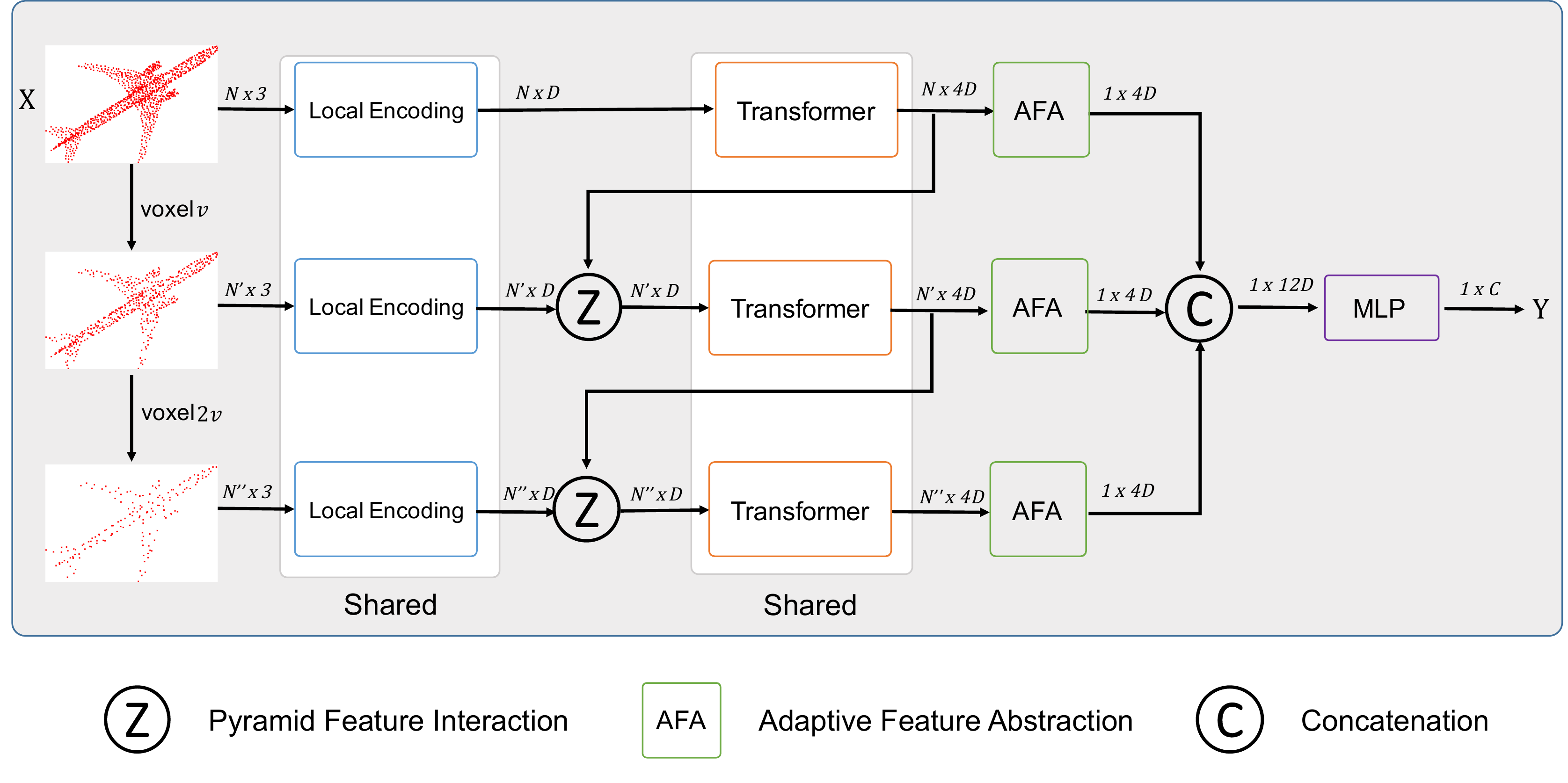}
\caption{The overall architecture of PV-Ada for point cloud classification. It takes original point cloud $\mathbf X$ as input and outputs $C$ confidence scores $\mathbf Y$. The modules of pyramid feature interaction and adaptive feature abstraction are detailed in \cref{subsec:encoder} and \cref{subsec:afa} respectively.}
\label{fig:pv-ada}
\end{figure*}

\paragraph{Pyramid feature interaction} As presented in \cref{fig:pv-ada}, we progressively voxelize the input point cloud $\mathbf{X}$ twice with voxel size $v$ and $2v$, denoting the voxelized point clouds as $\mathbf{X'} \in \mathbb R^{N' \times 3}$ and $\mathbf{X''} \in \mathbb R^{N'' \times 3}$. Pyramid feature interaction is conducted in two aspects. For one hand, local encoding and Transformer are weight-shared across $\mathbf X$, $\mathbf{X'} \in \mathbb R^{N' \times 3}$ and $\mathbf{X''} \in \mathbb R^{N'' \times 3}$. For the other hand, we devised feature interaction operator $z(\cdot, \cdot)$ to reuse learned fine-grained features. Three functions are considered: 
\begin{equation}
\begin{aligned}
  z_1(\mathbf F^1, \mathbf F^2) &= \mathbf F^2, \\
  z_2(\mathbf F^1, \mathbf F^2) &= h_1(\mathbf F^1) + \mathbf F^2, \\
  z_3(\mathbf F^1, \mathbf F^2) &= h_3(\text{cat}([h_2(\mathbf F^1), \mathbf F^2])), \\
  \label{eq:interaction}
\end{aligned}
\end{equation}
where $h_1(\cdot), h_2(\cdot)$ and $h_3(\cdot)$ are composed of Conv1d, normalization and LeakyReLU operations used for modifying feature dimensions. Finally, $\mathbf{F}_{\mathbf X'}^t \in \mathbb R^{N' \times {4D}}, \mathbf{F}_{\mathbf X''}^t \in \mathbb R^{N'' \times {4D}}$ are obtained in the same way as $\mathbf{F}_{\mathbf X}^t$. All of them are utilized in the following adaptive feature abstraction. 

\subsection{Adaptive feature abstraction} 
\label{subsec:afa}

Inspired by IA-SSD\cite{zhang2022not} and NgeNet\cite{zhu2022neighborhood}, we noticed that the idea of not all points being equal/important also applies to point cloud classification.

Adaptive feature abstraction takes $\mathbf{F}_{\mathbf X}^t \in \mathbb R^{N \times {4D}}$ as input and outputs point score $\mathbf{S}_{\mathbf X}^t \in \mathbb R^{N \times {1}}$ for the following pooling operation:
\begin{equation}
\begin{aligned}
  \mathbf{S}_{\mathbf X}^t &= g(\mathbf{F}_{\mathbf X}^t), \\
  \mathbf{F}_{\mathbf X} &= \text{MaxPooling}(\mathbf{S}_{\mathbf X}^t \cdot \mathbf{F}_{\mathbf X}^t),
  \label{eq:adaptive}
\end{aligned}
\end{equation}
where $g(\cdot)$ is implemented by Conv1d and $\mathbf{F}_{\mathbf X} \in \mathbb R^{4D}$. $\mathbf{F}_{\mathbf X'} \in \mathbb R^{4D}$ and $\mathbf{F}_{\mathbf X''} \in \mathbb R^{4D}$ are obtained in the same way, but learned parameters are not shared in the respective $g(\cdot)$.

We concat $\mathbf{F}_{\mathbf X}$, $\mathbf{F}_{\mathbf X'}$ and $\mathbf{F}_{\mathbf X''}$ as input, $C$ dimension score vector $\mathbf Y$ is outputted through MLPs.

\subsection{Loss}

Cross entropy loss with label smoothing is adopted in our model. Given predicted score $\mathbf Y \in \mathbb R^C$ and the ground truth one-hot label $\hat {\mathbf Y} \in \{0, 1\}^C$, loss function is defined as
\begin{equation}
\begin{aligned}
\hat {\mathbf y}^{soft}_i &=  \hat {\mathbf y}_i \cdot (1 - \alpha) + (1 - \hat {\mathbf y}_i) \cdot \alpha / (C - 1), \\
\mathcal L &= - \sum_{i = 0}^{C - 1} {\hat {\mathbf y}^{soft}_i \cdot \log \text{softmax}(\mathbf y_i)},
\end{aligned}
\end{equation}
where $\alpha$ is a label smoothing parameter.

\section{Experiments}
\label{sec:experi}

\subsection{Datasets}

\begin{figure*}[!t]
\centering
\includegraphics[width=.98\textwidth]{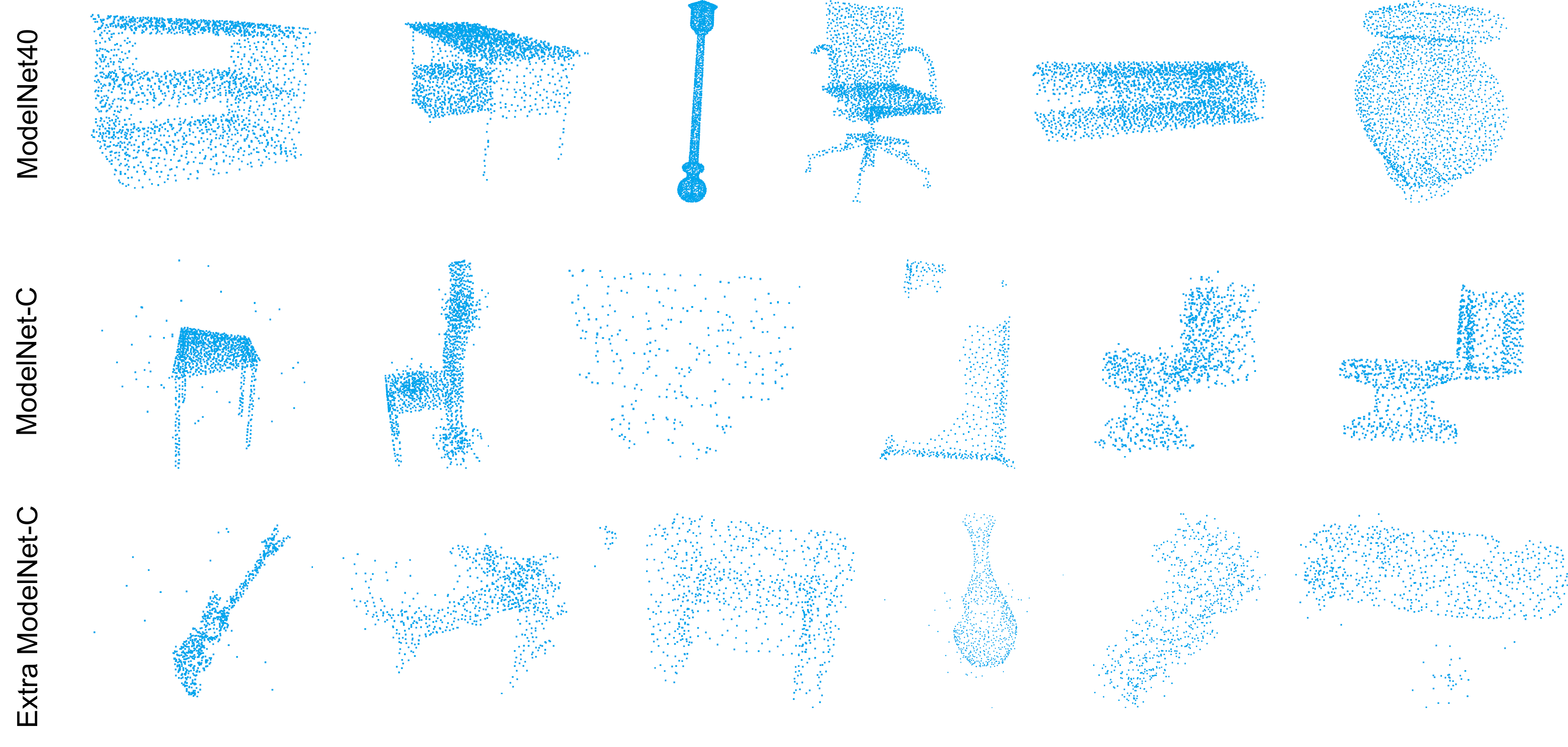}
\caption{Visualization of clean ModelNet40 \textit{testing set}, ModelNet-C (validation set) and Extrac ModelNet-C (testing set) is illustrated in the fisrt, second and third line, respectively. For the first and third line, point clouds are randomly selected. For the second line, point clouds are randomly selected from 6 corruption sets at severity level 5:  "Add Global", "Add Local", "Drop Global", "Drop Local", "Jitter" and "Scale".}
\label{fig:data}
\end{figure*}

\paragraph{Training set} We use ModelNet40\cite{wu20153d} \textit{training set} for model training. It collects 9,843 (of total 12, 311) CAD models from 40 man-made categories. In consistent with PointNet\cite{qi2017pointnet}, 1024 points are sampled from each aligned CAD model and then are normalized into a unit sphere.

\paragraph{Validation set} The proposed public ModelNet-C\cite{ren2022modelnet-c} is as our validation set, which corrupts ModelNet40 \textit{test set} with the aforementioned 7 atomic corruptions at 5 severity levels. In total, 35 corrupted sets are generated, with each contains 2468 point clouds. It's noted that points in different set may be different. For example, there are 524 and 1524 points in "Drop Local" set and "“Add Local" set at severity level 5, respectively.

\paragraph{Testing set} Test set is generated by a combination of different corruptions, rather than the separated ones in ModelNet-C, which is much more challenging. There are 24680 point clouds in total, and each contains 724 points. Some examples of training set, validation set and testing set are visualized in \cref{fig:data}.

\subsection{Evaluation metric}

Overall Accuracy (OA) or Corruption Error (CE) \cite{hendrycks2019benchmarking} are adopted to evaluate model performance on validation set and testing set. CE and mCE are calculated\cite{ren2022modelnet-c}:
\begin{equation}
\begin{aligned}
\text{CE}_i &= \frac{\sum_{l=1}^5 (1 - \text{OA}_{i,l})}{\sum_{l=1}^5(1 - \text{OA}_{i,l}^{\text{DGCNN}})}, \\
\text{mCE} &= \frac{1}{N}\sum_{i=1}^N\text{CE}_i,
\end{aligned}
\end{equation}
where $\text{OA}_{i, l}$ is the overall accuracy on corrupted set using corruption type $i$ at severity level $l$, $\text{OA}_{i,l}^{\text{DGCNN}}$ is the baseline overall accuracy using DGCNN\cite{wang2019dynamic}, and $N=7$ is the number of atomic corruption types.

\subsection{Implementation details}

PV-Ada is implemented with PyTorch\cite{paszke2019pytorch}. We train the model 350 epochs on a single RTX3090 GPU card with batch size 64. SGD is used with initial learning rate 0.1, momentum 0.9 and weight decay 0.0005. Learning rate is scheduled using a cosine annealing with $T_{\max}=350$ and $\eta_{\min}=0.001$. Voxel size $v$ is set to 0.05, $k$ in local encoding is set to 32, feature dimension $D$ is set to 128, and label smoothing parameter $\alpha$ is set to 0.2.

\paragraph{Data augmentation} In consistent with \cite{ren2022modelnet-c}, three augmentations are utilized in our model: random anisotropic scaling in the range $[\frac{2}{3}, \frac{3}{2}]$, random translation in the range $[-0.2, 0.2]$ and WOLFMix. WOLFMix is designed based on PointWOLF\cite{kim2021point} and RSMix\cite{lee2021regularization}. Different from RPC\cite{ren2022modelnet-c}, which leverages RSMix augmentation in probability in each iteration, we always leverage RSMix augmentation for the final 50 epochs during training. Besides, we explore other data augmentation, such as Tapering\cite{perez20223deformrs}, in our experiments.

\subsection{Comparison to the state-of-the-arts}

Results for overall accuracy on testing set (ExtraOA), validation set (mOA) and clean ModelNet40 \textit{testing set} are reported in \cref{tab:acc}. Equipped with WOLFMix, we achieve 0.860 ExtraOA, improving 6.3\% compared with the state-of-the-art methods. We achieve 0.884 mOA on the validation set, outperforming all the published methods. Detailed results on the 7 corruption sets are also reported. We improve accuracy on "Jitter" set compared to PCT, GDANet and RPC, while the performance on other corruption sets does not deteriorate significantly. We achieve 0.923 on the clean ModelNet-40 \textit{testing set}. Though droping about 1\% compared to the SOTA models, a relative lower baseline is adopted as our model(see \cref{tab:aba_pos}). With Tapering augmentation, we achieve 0.865 ExtraOA on the testing set. Results for mCE are reported in \cref{tab:mce}, where we achieve the lowest mCE on validation set.

\begin{table*}
\setlength{\tabcolsep}{1pt}
\centering\scriptsize
\caption{Full results for Overall Accuracy (OA) on testing set (ExtraOA), validation set (mOA) and clean ModelNet40 \textit{testing set}. W.M. is short for WOLFMix and Tp is short for Tapering. Bold: best in column. Underline: second best in column.}
\label{tab:acc}
\begin{tabular}{lc|c|c|ccccccc}
\toprule
{} & ExtraOA $\uparrow$ & Clean $\uparrow$ & mOA$\uparrow$ & Scale & Jitter & Drop-G & Drop-L & Add-G & Add-L & Rotate \\
\midrule
PointNet\cite{qi2017pointnet}+W.M. & - & 0.884 & 0.743 & 0.801 & \textbf{0.850} & 0.857 & 0.776 & 0.343 & 0.807 & 0.768 \\
PCT\cite{guo2021pct}+W.M. & - & \textbf{0.934} & 0.873 & 0.906 & 0.730 & \textbf{0.906} & \textbf{0.898} & 0.912 & 0.861 & 0.895 \\
GDANet\cite{xu2021learning}+W.M. & 0.797 & \textbf{0.934} & 0.871 & \textbf{0.915} & 0.721 & 0.868 & \underline{0.886} & 0.910 & 0.886 & \textbf{0.912} \\
RPC\cite{ren2022modelnet-c}+W.M. & 0.739 & \underline{0.933} & 0.865 & 0.905 & 0.694 & 0.895 & \underline{0.894} & 0.902 & 0.868 & \underline{0.897} \\
\midrule
PV-Ada+W.M. & \underline{0.860} & 0.923 & \textbf{0.884} & \underline{0.911} & \underline{0.796} & \underline{0.9} & 0.88 & \textbf{0.915} & \textbf{0.89} & 0.896 \\
PV-Ada+W.M.+Tp & \textbf{0.865} & 0.911 & \underline{0.88} & 0.907 & 0.792 & 0.897 & 0.874 & \underline{0.914} & 0.884 & 0.892 \\
\bottomrule
\end{tabular}
\end{table*}

\begin{table*}
\setlength{\tabcolsep}{1pt}
\centering\footnotesize
\caption{Full results for mCE on validation set. Bold: best in column. Underline: second best in column.}
\label{tab:mce}
\begin{tabular}{l|c|ccccccc}
\toprule
{} & \textbf{mCE}  $\downarrow$& Scale & Jitter & Drop-G & Drop-L & Add-G & Add-L & Rotate \\
\midrule
DGCNN\cite{wang2019dynamic}+W.M. & 0.590 & 0.989 & 0.715 & 0.698 & 0.575 & \textbf{0.285} & \underline{0.415} &  \underline{0.451}\\
PCT\cite{guo2021pct}+W.M. & 0.574 & 1.000 & 0.854 & \textbf{0.379} & \textbf{0.493} & 0.298 & 0.505 & 0.488 \\
GDANet\cite{xu2021learning}+W.M. &  \underline{0.571} & \textbf{0.904} & 0.883 & 0.532 & 0.551 & 0.305 & \underline{0.415} & \textbf{0.409} \\
RPC\cite{ren2022modelnet-c}+W.M. & 0.601 & 1.011 & 0.968 & 0.423 & \underline{0.512} & 0.332 & 0.480 & 0.479 \\
\midrule
PV-Ada+W.M. & \textbf{0.538} & \underline{0.947} & \textbf{0.652} & \underline{0.403} & 0.58 & \underline{0.292} & \textbf{0.4} & 0.493 \\
\bottomrule
\end{tabular}
\end{table*}

\subsection{Ablation studies}

Ablation studies on the proposed modules are summarized in \cref{tab:aba_pos}.
\paragraph{Base} One classification model whose backbone is composed of simple local encoding and Transformer is adopted as the baseline model. 
\paragraph{P-V} With Point-Voxel architecture whose local encoding and Transformer shared, the accuracy on validation set is improved by 1.7\%. 
\paragraph{Best val} We choose the weight with best performance on validation set, instead of on clean ModelNet40 \textit{testing set}. With this trick, we achieve 0.848 ExtraOA, improving 0.5\% on testing set.
\paragraph{RSMix} RSMix is always leveraged in the final 50 epochs during training. With this trick, we achieved 0.855 ExtraOA on testing set.
\paragraph{AFA} Adaptive feature abstraction is the proposed module introduced in \ref{subsec:afa}. With AFA module, we achieve 0.860 ExtraOA on testing set.
\paragraph{Tapering} With Tapering augmentation, we achieve the best ExtraOA 0.865 on testing set. However, performance on both validation set and clean ModelNet40 `testing set` drops. Also, we observed that training with Tapering is not stable. We didn't conduct more experiments on Tapering.

Except the above proposed modules and positive tricks, we also explore other module designs. The results are summarized in \cref{tab:aba_neg}.

\begin{table*}[!t]
\setlength{\tabcolsep}{5pt}
\centering\small
\caption{Ablation studies on the proposed modules and positive tricks on ModelNet-C.}
\label{tab:aba_pos}
\begin{tabular}{cccccc|ccc}
\toprule
Base & P-V & Best val & RSMix & AFA & Tapering & clean $\uparrow$ & mOA $\uparrow$ & ExtraOA $\uparrow$ \\
\midrule
\cmark & & & & & & 0.928 & 0.862 & - \\
\cmark & \cmark & & & & & 0.927 & 0.879 & 0.843 \\ 
\cmark & \cmark & \cmark & & & & 0.923 & 0.881 & 0.848 \\
\cmark & \cmark & \cmark & \cmark & & & 0.921 & 0.882 & 0.855 \\
\cmark & \cmark & \cmark & \cmark & \cmark & & 0.923 & 0.884 & 0.860 \\
\cmark & \cmark & \cmark & \cmark & \cmark & \cmark & 0.911 & 0.88 & 0.865 \\

\bottomrule
\end{tabular}
\end{table*}

\begin{table}[!t]
\caption{Ablation studies on the negative designed modules.}
\setlength{\tabcolsep}{26pt}
\centering
\label{tab:aba_neg}
\begin{tabular}{l|cc}
\toprule
Model & clean $\uparrow$ & mOA $\uparrow$ \\ 
\midrule
Base + P-V ($z_1$) & 0.927 & 0.879 \\ 
Base + P-V ($z_2$) & 0.932 & 0.862 \\
Base + P-V ($z_3$) & 0.926 & 0.868 \\
\midrule
Base + P-V (share) & 0.927 & 0.879 \\
Base + P-V (w.o. share) & 0.929 & 0.855 \\
\midrule
Base + P-V (\#2 voxel) & 0.927 & 0.879 \\
Base + P-V (\#3 voxel) & 0.914 & 0.868 \\
\bottomrule
\end{tabular}
\end{table}

\paragraph{Pyramid feature interaction} We compare the performance between $z_1(\cdot, \cdot)$, $z_2(\cdot, \cdot)$ and $z_3(\cdot, \cdot)$. As shown the top-3 lines in \cref{tab:aba_neg}, the highest mOA 0.879 is achieved with $z_1(\cdot, \cdot)$. It indicates that better performance may be obtained if there is no feature interaction.
\paragraph{Shared weights} Higher mOA is obtained when the weights are shared by local encoding and Transformer modules. Meanwhile, less parameters are needed when the weights are shared in some modules.
\paragraph{The times of iterative voxelization} We increase the iterative voxlization times to 3 and achieve the worse mOA 0.868. The reason for this maybe the voxel size ($4v$) is too large for ModelNet and ModelNet-C.

\subsection{Model size and inference time}

The model size and inference time are reported in \cref{tab:model_size}. There are about 3.16M parameters in PV-Ada. Inference time is averaged on testing set using batch size 1 to predict. 

\begin{table}[!t]
\caption{Model size and inference time}
\setlength{\tabcolsep}{28pt}
\centering
\label{tab:model_size}
\begin{tabular}{l|cc}
\toprule
Model & \# parameters & inference time \\ 
\midrule
PV-Ada & 3.16 M & 15.7 ms \\
\bottomrule
\end{tabular}
\end{table}

\section{Discussion and conclusion}
\label{sec:conc}

Equipped with a stronger backbone, better performance may be achieved in point cloud classification under corruptions. Data augmentations, such as Tapering, Twisting and Shearing are not fully explored in our experiments, and we are not sure whether they are positive for model training.

In this report, we propose PV-Ada for robust point cloud classification under corruptions. Point-voxel encoder with shared local encoding and Transformer is introduced for point feature extraction. Then adaptive feature abstraction is proposed for point cloud representation. With several tricks, our PV-Ada ranks the second place in ModelNet-C classification track of PointCloud-C Challenge 2022, with ExtraOA being 0.865. Additional, PV-Ada outperforms the published state-of-the-art motheds on ModelNet-C dataset.

\clearpage
%
%
\bibliographystyle{splncs04}
\bibliography{egbib}
\end{document}